\documentclass[letterpaper]{article} % DO NOT CHANGE THIS
\usepackage{aaai2026}  % DO NOT CHANGE THIS
\usepackage{times}  % DO NOT CHANGE THIS
\usepackage{helvet}  % DO NOT CHANGE THIS
\usepackage{courier}  % DO NOT CHANGE THIS
\usepackage[hyphens]{url}  % DO NOT CHANGE THIS
\usepackage{graphicx} % DO NOT CHANGE THIS
\usepackage{natbib}  % DO NOT CHANGE THIS AND DO NOT ADD ANY OPTIONS TO IT
\usepackage{caption} % DO NOT CHANGE THIS AND DO NOT ADD ANY OPTIONS TO IT
\usepackage{algorithm}
\usepackage{algorithmic}
\usepackage{amsmath}
\usepackage{booktabs} 
\usepackage{multirow}
\usepackage{newfloat}
\usepackage{listings}
\DeclareCaptionStyle{ruled}{labelfont=normalfont,labelsep=colon,strut=off} % DO NOT CHANGE THIS
\floatstyle{ruled}
\newfloat{listing}{tb}{lst}{}
\floatname{listing}{Listing}
\title{SOPSeg: Prompt-based Small Object Instance Segmentation in Remote Sensing Imagery}
\author{
    Chenhao Wang\textsuperscript{\rm 1,2}\equalcontrib,
    Yingrui Ji\textsuperscript{\rm 1,2}\equalcontrib,
    Yu Meng\textsuperscript{\rm 1},
    Yunjian Zhang\textsuperscript{\rm 3}, 
    Yao Zhu\textsuperscript{\rm 4},
}
\affiliations{
    \textsuperscript{\rm 1} Aerospace Information Research Institute, Chinese Academy of Sciences \\
    \textsuperscript{\rm 2} School of Electronic, Electrical and Communication Engineering, University of Chinese Academy of Sciences \\
    \textsuperscript{\rm 3} Institution of Information Engineering, Chinese Academic of Sciences
    \textsuperscript{\rm 4} Zhejiang University, Hangzhou

}

\usepackage{bibentry}
\begin{document}

\maketitle

\begin{abstract}
Extracting small objects from remote sensing imagery plays a vital role in various applications, including urban planning, environmental monitoring, and disaster management. While current research primarily focuses on small object detection, instance segmentation for small objects remains underexplored, with no dedicated datasets available. This gap stems from the technical challenges and high costs of pixel-level annotation for small objects.
While the Segment Anything Model (SAM) demonstrates impressive zero-shot generalization, its performance on small-object segmentation deteriorates significantly, largely due to the coarse 1/16 feature resolution that causes severe loss of fine spatial details.
To this end, we propose SOPSeg, a prompt-based framework specifically designed for small object segmentation in remote sensing imagery. It incorporates a region-adaptive magnification strategy to preserve fine-grained details, and employs a customized decoder that integrates edge prediction and progressive refinement for accurate boundary delineation. Moreover, we introduce a novel prompting mechanism tailored to the oriented bounding boxes widely adopted in remote sensing applications.
% Extensive experimental results demonstrate that SOPSeg outperforms existing methods in small object segmentation scenarios. By facilitating efficient dataset construction for remote sensing tasks, it enables us to contribute a comprehensive small object instance segmentation dataset. We will release both the model and the dataset to support future research.
SOPSeg outperforms existing methods in small object segmentation and facilitates efficient dataset construction for remote sensing tasks. We further construct a comprehensive small object instance segmentation dataset based on SODA-A, and will release both the model and dataset to support future research.
\end{abstract}


\section{Introduction}
Remote sensing imagery plays a critical role in a wide range of real-world applications, including urban planning, environmental monitoring, and precision agriculture. Among the targets of interest in these applications, small objects such as vehicles, plane, and ships typically occupy no more than 32×32 pixels in high-resolution imagery, yet they convey essential semantic and operational information for downstream tasks. Consequently, accurately extracting small objects is of great importance, but remains a highly challenging task due to their limited size and complex visual characteristics.

Benchmarks such as SODA-A \cite{gong2023soda} have significantly advanced small object detection in remote sensing imagery. However, they provide only bounding box annotations, which constrain models to coarse localization and fail to capture precise object shapes. Consequently, most existing  works focus on object detection rather than instance segmentation, limiting fine-grained scene understanding.

Instance segmentation for small objects remains largely underexplored, primarily due to the lack of suitable datasets. Constructing such datasets is highly labor-intensive, error-prone, and requires substantial domain expertise. Although the Segment Anything Model (SAM) \cite{kirillov2023segment}, trained on over one billion masks, demonstrates strong zero-shot generalization capabilities, its direct application to high-resolution remote sensing imagery leads to notable performance degradation for small objects. We attribute this limitation to the architectural design of SAM: its vision transformer encoder downsamples input images to 1/16 of the original resolution to reduce computational cost. While effective for typical object sizes, this aggressive downsampling results in the loss of fine-grained details that are critical for accurately identifying small targets.

To this end, we propose \textbf{SOPSeg} (\textbf{S}mall \textbf{O}bject \textbf{P}rompted \textbf{Seg}mentation), a novel framework that adapts SAM for robust small-object instance segmentation in remote sensing imagery. Our approach introduces three key innovations: (1) Region-adaptive magnification, which adaptively crops and resizes object regions to preserve fine details lost in downsampling, enabling accurate segmentation of small instances with minimal overhead;
(2) An edge-aware decoder, which integrates boundary prediction and progressive multi-scale refinement to produce sharper and more accurate object masks;
(3) An oriented prompting mechanism, which enables the use of rotated bounding boxes common in aerial imagery, improving SAM’s ability to handle objects at arbitrary orientations.

We train and validate SOPSeg on the iSAID dataset \cite{Zamir2019isaid}, selecting 7 out of 15 categories that best represent small object challenges in remote sensing imagery. Generalization ability is further evaluated on the NWPU-VHR10 \cite{su2019object} and SAT-MTB \cite{li2023multitask} benchmarks. Experimental results show that SOPSeg significantly outperforms the original SAM and other prompt-based segmentation methods across all datasets.
%Our method achieves markedly higher mask IoU for the smallest object instances, validating the effectiveness of our region magnification and edge refinement strategies.

\begin{figure*}[htbp]
    \centering
    \includegraphics[width=0.8\textwidth]{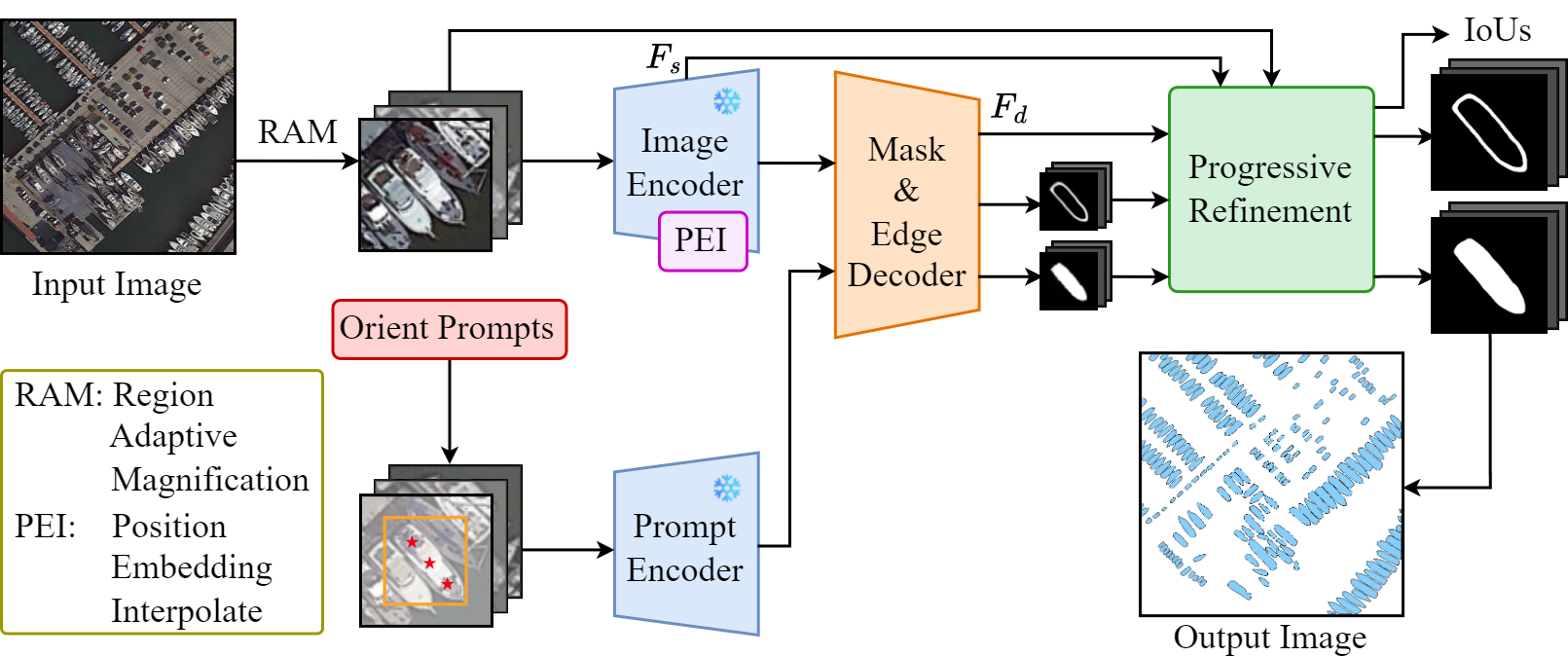}
    \caption{Overview of the SOPSeg Framework. The input remote sensing image is first cropped and resized into multiple high-resolution patches of uniform size. These patches are then fed into the SAM image encoder, where Position Embedding Interpolation (PEI) is applied to support arbitrary input sizes. An oriented prompt, consisting of a horizontal bounding box and three keypoints aligned with the orientation of the object, is encoded via the SAM prompt encoder to guide segmentation. Coarse masks and edges are generated and refined progressively to yield accurate high-resolution segmentation. $\mathbf{F_d}$ and $\mathbf{F_s}$ denote deep image features from decoder and shallow image features from image encoder.}
    \label{fig:architecture}
\end{figure*}

To demonstrate its practical utility, we further apply SOPSeg to assist in constructing a small object instance segmentation dataset. Specifically, we automatically generate approximately 709k instance masks for small objects based on images and oriented bounding boxes from the SODA-A dataset, followed by manual filtering to remove a small number of abnormal annotations. The resulting dataset \textbf{ReSOS} (\textbf{Re}mote Sensing \textbf{S}mall \textbf{O}bject \textbf{S}egmentation), represents the first large-scale instance segmentation benchmark focused on small objects in remote sensing imagery. We plan to publicly release both the model and the dataset to provide training and evaluation resources for future research on small object analysis.

In summary, our contributions are threefold: 
\begin{itemize}
\item We propose SOPSeg, a prompt-based framework that adapts SAM for small object instance segmentation, integrating region-adaptive magnification and edge-aware refinement decoding to enhance mask accuracy.
\item We develop an oriented prompting mechanism enabling accurate segmentation of objects at arbitrary orientations.
\item We construct and release ReSOS dataset, the first large-scale instance segmentation dataset specifically designed for small objects in remote sensing. It contains pixel-level annotations for over 709k instances and aims to support future research.
\end{itemize}

\section{Related Work}
\textbf{Small Object Detection and Segmentation in Remote Sensing.} Small object analysis in remote sensing has attracted significant research interest due to its practical importance. Early detection methods relied on hand-crafted features and traditional machine learning classifiers \cite{cheng2016survey}. However, these approaches struggled with the complex backgrounds and varying scales characteristic of aerial imagery. Ding \cite{ding2019learning} proposed a rotation-invariant detector specifically designed for aerial images, while Yang \cite{yang2019scrdet} introduced SCRDet to handle the multi-scale and multi-orientation challenges. Recent methods have focused on feature enhancement strategies. RMSIN \cite{liu2024RMSIN} employs interaction modules to effectively capture complex spatial scales and orientations for accurate segmentation in remote sensing imagery. For instance, FCOS-RS \cite{li2020feature} adapts the anchor-free FCOS detector for remote sensing by incorporating multi-scale feature fusion. Similarly, Oriented R-CNN \cite{xie2021oriented} extends Faster R-CNN \cite{ren2016faster} with oriented region proposals to better capture arbitrarily oriented objects.
Despite progress in detection, instance segmentation of small objects remains largely unexplored. The few existing works primarily focus on specific object categories. Zhang \cite{zhang2017complex} developed a ship instance segmentation method using polar coordinates, while Zhao \cite{zhao2021building} proposed building extraction techniques. UGBS \cite{yang2024Exploring} explored interactive user guidance mechanisms to achieve more accurate building segmentation from high-resolution remote sensing images, demonstrating the potential of human-in-the-loop approaches.  However, these category-specific approaches do not generalize to diverse small objects. The scarcity of segmentation methods stems from the lack of appropriate datasets and the inherent difficulty of obtaining pixel-level annotations for tiny objects.
% The challenge is further compounded by the unique characteristics of remote sensing imagery. Small objects often exhibit low contrast against complex backgrounds, suffer from atmospheric effects, and appear at arbitrary orientations. Traditional segmentation networks like Mask R-CNN \cite{he2017mask} show significant performance degradation when directly applied to small remote sensing objects, primarily due to insufficient feature representation at coarse resolutions. This limitation motivates the need for specialized architectures that preserve fine-grained details throughout the processing pipeline.

\noindent
\textbf{Segment Anything Model and Applications.} The Segment Anything Model (SAM) \cite{kirillov2023segment} represents a paradigm shift in image segmentation through its foundation model approach. SAM's versatility stems from its flexible prompting mechanism. Users can specify objects of interest through points, bounding boxes, or coarse masks, enabling interactive segmentation workflows. 
Recent works have explored SAM's potential in remote sensing applications. SAMRS \cite{wang2023samrs} leverages SAM to automatically convert object detection datasets into instance segmentation datasets, demonstrating its utility for large-scale annotation tasks. SAM2 \cite{ravi2024sam2segmentimages} enhanced segmentation accuracy on both images and videos. RSPrompter \cite{chen2023rsprompter} introduces auxiliary prompts specifically designed for remote sensing imagery to improve SAM's performance. SAM-Adapter \cite{chen2023sam} proposes lightweight adapters to adapt SAM for domain-specific tasks while preserving its zero-shot capabilities. ROS-SAM \cite{shan2025ros} specifically targets moving object segmentation in remote sensing videos by leveraging LoRA-based adaptation and a context-aware decoder. It primarily focuses on objects with sufficient motion patterns rather than addressing the challenges of small object segmentation. HQ-SAM \cite{ke2023hqsegment} addresses the issue of coarse mask boundaries in the original SAM by introducing a learnable High-Quality Output Token. Matting Anything \cite{Li2024matting} extends SAM to the image matting task by predicting precise alpha channels for objects with complex boundaries. Nevertheless, constrained by the resolution of low-level features, existing approaches exhibit limited performance in small object segmentation — a key challenge that this study seeks to systematically tackle.
% However, several studies have identified limitations when applying SAM to remote sensing imagery, particularly for small objects.  It comprises fine-grained details for segmenting small objects.

\section{Methodology}
To bridge the gap between generic segmentation models and the unique demands of small object segmentation in remote sensing, we propose SOPSeg, a prompt-based framework that introduces three key improvements over SAM: a region-adaptive magnification strategy, an oriented prompt mechanism, and an enhanced decoder with integrated edge prediction. The overall architecture is illustrated in Figure \ref{fig:architecture}.

% \noindent
\subsection{Region-Adaptive Magnification Strategy}
The core challenge in small object segmentation lies in preserving spatial details during feature extraction. SAM's vanilla image encoder processes images at a fixed resolution, downsampling features to 1/16 of the original size. For small objects occupying only $32 \times 32$ pixels, this results in feature representations of merely $2 \times 2 $ pixels, causing severe information loss.

\begin{figure}[htbp]
\centering
\includegraphics[width=0.47\textwidth]{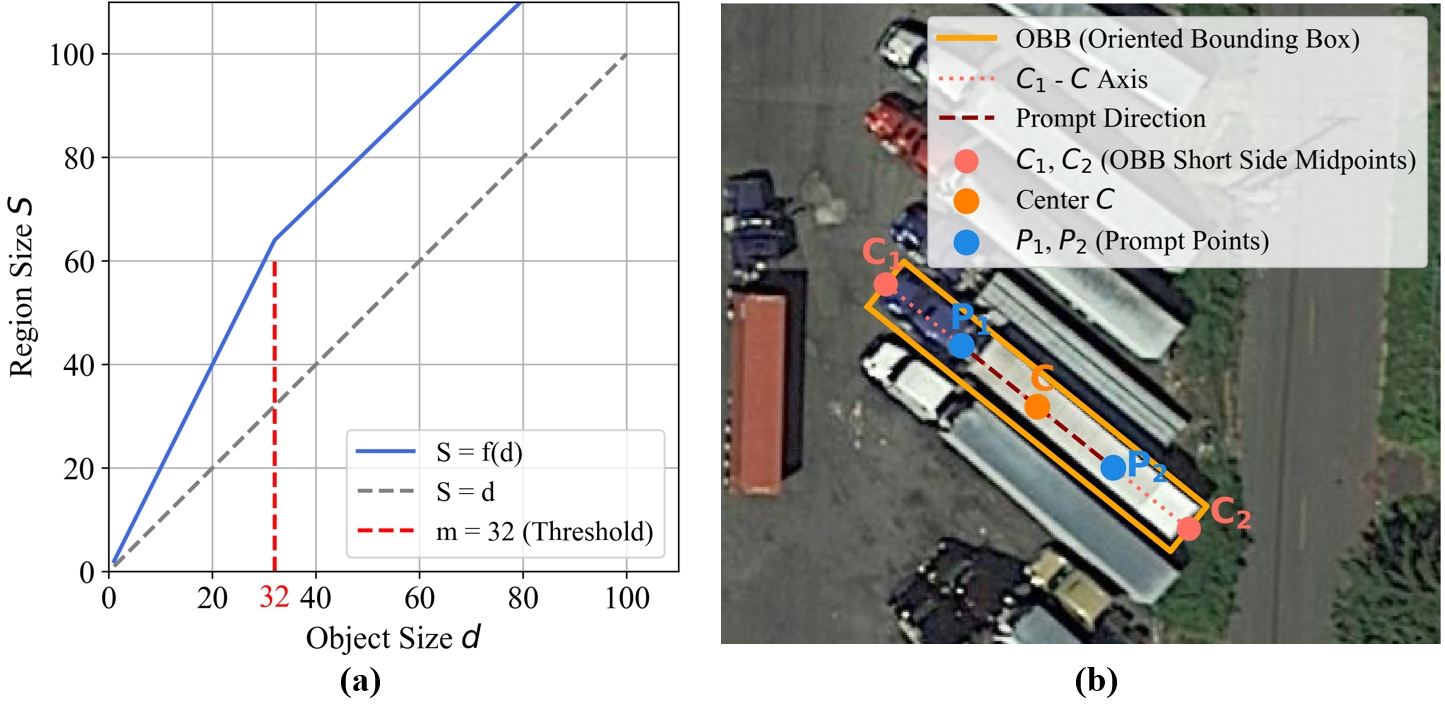}
\caption{(a) The figure illustrates the relationship between the extracted region size $S$ and object dimension $d = max(w, h)$. (b) Illustration of the oriented prompting mechanism for rotated objects. The final prompt points include $P_1$, $C$, and $P_2$.}
\label{fig:merge}
\end{figure}

Our region-adaptive magnification strategy addresses this limitation through adaptive region extraction and resizing. Given an input image and the bounding box $b = (x, y, w, h)$ of a instance, we extract a square local region with boundaries $(x_s, y_s, S, S)$, where the size $S$ is determined by the object size $d = \max(w, h)$. Each region is then resized to a fixed resolution of $S_{\text{in}} \times S_{\text{in}}$ before being fed into the model. The relationship between $S$ and $d$ is formulated as:
\begin{equation}
S = \begin{cases}
k_0 \cdot d, & \text{if } d < m \\
k \cdot d + (k_0 - k) \cdot m, & \text{if } d \geq m
\end{cases}
\end{equation}
where:
$k = \frac{S_{\text{max}} - k_0 \cdot m}{S_{\text{max}} - m}$ and satisfies the boundary condition $S = S_{\text{max}}$ when $d = S_{\text{max}}$. A visual illustration is provided in Fig.~\ref{fig:merge}(a).
We empirically set $m = 32$, $k_0 = 2$, and $S_{\text{max}} = 1024$. Here, $k_0$ represents the initial region expand factor for objects smaller than the threshold $m$. Since all regions are resized to a fixed input size $S_{\text{in}}$, the object magnification is $S_{\text{in}}/S$. Smaller $S$ yields larger magnification, which benefits small objects by enhancing fine details. For objects larger than the threshold, appropriately reducing surrounding context increases their magnification while still retaining essential context. This adaptive design balances detail preservation for small instances and contextual integrity for large ones. 

We compute the top-left coordinates $(x_s, y_s)$ based on the object's bounding box center and the desired region size:
% \[
% x_s = x - a_x(S - w), \quad y_s = y - a_y(S - h)
% \]
\begin{equation}
\begin{bmatrix}
x_s \\
y_s
\end{bmatrix}
=
\begin{bmatrix}
x \\
y
\end{bmatrix}
-
\begin{bmatrix}
a_x (S - w) \\
a_y (S - h)
\end{bmatrix}
\end{equation}
where $a_x$ and $a_y$ control the object's position within the extracted region. During training, we set $a_x, a_y \in [0.3, 0.7]$ randomly to improve generalization capability, ensuring objects appear at various positions rather than always centered.

 SAM's default $1024 \times 1024$ input resolution is designed for processing entire images containing objects of various sizes. However, when focusing on small objects through region extraction, this high resolution becomes computationally wasteful—most pixels represent irrelevant background rather than the target object. Thus we set the input size to $S_{\text{in}} = 256$, which preserves sufficient detail for accurate segmentation while significantly reducing computational overhead. For instance, if a small object originally spans $32 \times 32$ pixels, the extracted region size $S = 64$, and the object is effectively magnified by a factor of $S_{\text{in}} / S = 4$.

\noindent
\textbf{Position Embedding Interpolation.}
Since SAM's positional embedding weights are input-size dependent, the pretrained embeddings trained on $1024 \times 1024$ inputs cannot be directly reused. We address this through bilinear interpolation:

\begin{equation}
    \text{PE}_{\text{target}} = \text{Interpolate}(\text{PE}_{1024}, S_{\text{in}}, S_{\text{in}})
\end{equation}

where $S_{\text{in}} = 256$ for small object processing. This interpolation preserves the relative spatial patterns while adapting to the new resolution.

The combination of region extraction, magnification, and reduced input resolution creates an efficient pipeline: small objects are first magnified to an adequate size, then processed at lower input size without losing critical details.

\subsection{Oriented Prompt Mechanism}

Remote sensing objects frequently appear at arbitrary orientations, posing challenges for standard segmentation models. To this end, we propose a strategy that encodes object orientation using strategically placed points, thereby enabling the original SAM—designed for axis-aligned bounding boxes—to effectively handle rotated objects in aerial imagery, all without requiring any modifications to its architecture.

For each oriented bounding box, we extract three key geometric points: the geometric center $C$, and the midpoints $C_1$ and $C_2$ of the two shorter sides. The line segment $\overline{C_1C_2}$ naturally defines the object's principal axis. Apart from $C$, we generate two prompt points along the principal axis:
\begin{equation}
    P_1 = \frac{C + C_1}{2} ,\quad P_2 = \frac{C + C_2}{2}
\end{equation}

Fig. \ref{fig:merge}(b) illustrates our oriented prompt mechanism on a real example from aerial imagery.

These points encode both spatial and directional information:
\begin{itemize}
    \item The vector $\overrightarrow{P_1P_2}$ implicitly represents the object's orientation.
    \item The distance $||P_1 - P_2||$ correlates with the object's length along its principal axis.
    \item All points remain well within object boundaries, ensuring reliable prompting.
\end{itemize}

The three points $(P_1, C, P_2)$ are directly processed through SAM's pretrained point encoder:
\begin{equation}
    E_{\text{points}} = \text{PointEncoder}([P_1, C, P_2])
\end{equation}

Combined with the horizontal bounding box prompt, this provides comprehensive spatial guidance:
\begin{equation}
    E_{\text{prompt}} = \text{Concat}(\text{BoxEncoder}(b_{\text{horizontal}}), E_{\text{points}})
\end{equation}

This design maintains full compatibility with SAM's pretrained weights while effectively handling arbitrary orientations. The approach is particularly well-suited for elongated objects prevalent in remote sensing, such as vehicles and ships, where the short-edge midpoints naturally capture the object's dominant direction. 
%By encoding orientation through point positions rather than explicit rotation parameters, we avoid introducing new learnable components while achieving robust orientation-aware segmentation.

\subsection{Enhanced Decoder with Edge Prediction}
Despite the region magnification strategy, small objects in remote sensing imagery still suffer from boundary ambiguity due to complex backgrounds. 
%The vanilla SAM decoder, while effective for regular objects, often produces blurred edges when dealing with features that have been downsampled.
We introduce an auxiliary edge prediction path and progressive refinement, enhancing fine-grained delineation of small instances.

\noindent
\textbf{Stage 1: Parallel Edge Prediction.} We augment the SAM decoder with a learnable edge token $\mathbf{T}_\text{edge}$, which collaborates with the original mask tokens $\mathbf{T}_\text{mask}$ to capture boundary-specific information. The  $\mathbf{T}_\text{edge}$, $\mathbf{T}_\text{mask}$, and prompt tokens perform bidirectional attention with the image features, resulting in updated representations: $\mathbf{T}_\text{edge}^{(1)}$, $\mathbf{T}_\text{mask}^{(1)}$, and $\mathbf{F}_d$. These are then used to generate two parallel outputs:

\begin{align}
\mathbf{M}_0 &= \text{MLP}_{\text{mask}}(\mathbf{T}_\text{mask}^{(1)}) \cdot O_{\text{mask}}(\mathbf{F}_d)\\
\mathbf{E}_0 &= \text{MLP}_{\text{edge}}(\mathbf{T}_\text{edge}^{(1)}) \cdot O_{\text{edge}}(\mathbf{F}_d)
\end{align}

Here, $\text{MLP}_{\text{edge}}$ and $O_{\text{edge}}$ follow the same architectural design as the mask prediction modules in SAM. $\mathbf{M}_0$ and $\mathbf{E}_0$ denote the initial mask and edge predictions at a resolution of 1/4 the input size.

\noindent
\textbf{Stage 2: Progressive Refinement.} Initial predictions capture basic structure but lack fine details critical for small objects. We employ multi-scale refinement that gradually improves both masks and edges through iterative processing.
The refinement takes four inputs: deep image features $\mathbf{F}_{d}$ after attention from the decoder, shallow features $\mathbf{F}_{s}$ from the image encoder, the original image $\mathbf{I}$, and the initial predictions $\mathbf{P}_0 = [\mathbf{M}_0; \mathbf{E}_0]$ from Stage 1.

% , and combined predictions $\mathbf{P}_4 = [\mathbf{M}_0; \mathbf{E}_0]$. The subscript 4 indicates the downsampling factor relative to the original image.

Both the shallow and deep image features first undergo $2 \times$ upsampling and channel dimension reduction mapping for efficient processing. The shallow features $\mathbf{F}_{16s}$ ($\mathbf{F}_{s}$) are processed through convolution, normalization, and $2 \times$ upsampling to produce $\mathbf{F}_{8s}$. Similarly, decoder features $\mathbf{F}_{16d}$ ($\mathbf{F}_{d}$) are mapped and upscaled to $\mathbf{F}_{8d}$.

\noindent
\textbf{Multi-Scale Refinement.} The refinement operates across three spatial scales: $1/8 \rightarrow 1/4 \rightarrow 1/2 \rightarrow 1/1$, progressively enhancing both mask and edge predictions.

\begin{itemize}
    \item \textbf{Scale 1/8 to 1/4.} We concatenate the upsampled features $\mathbf{F}_{8s}$ and $\mathbf{F}_{8d}$ with the downsampled image $\mathbf{I}_8$ and predictions $\mathbf{P}_8$, and pass them through the first residual refinement block $\mathcal{R}_1$:
    \begin{equation}
        \mathbf{X}_4 = \mathcal{R}_1([\mathbf{F}_{8s}; \mathbf{F}_{8d}; \mathbf{I}_8; \mathbf{P}_8])
    \end{equation}
    The refined feature $\mathbf{X}_4$ is then mapped to updated predictions $\mathbf{P}_4 = [\mathbf{M}_4; \mathbf{E}_4]$ via the output head $\phi_1$:
    \begin{equation}
        \mathbf{P}_4 = \phi_1(\mathbf{X}_4)
    \end{equation}

    \item \textbf{Scales 1/4 to 1/2 to 1/1.} We apply the same refinement pattern iteratively. At each scale $i \in \{4, 2\}$, we use a residual refinement block $\mathcal{R}_j$ and an output head $\phi_j$ to generate updated predictions:
    \begin{equation}
        \mathbf{X}_{i/2} = \mathcal{R}_{j}([\mathbf{X}_i; \mathbf{I}_i; \mathbf{P}_i]), \quad \mathbf{P}_{i/2} = \phi_j(\mathbf{X}_{i/2})
    \end{equation}
    where $\mathbf{I}_i$ is the image and prediction maps downsampled to resolution $1/i$, while $\mathbf{P}_i$ is the output from last iteration.

    \item \textbf{IoU Prediction.} At the final stage, the refined feature $\mathbf{X}_1$ is also used to predict the mask quality score $p_{\text{iou}}$ via a lightweight head consisting of a convolutional layer, ReLU activation, adaptive average pooling, and linear projection:
    \begin{equation}
        p_{\text{iou}} = \text{IoU}(\mathbf{X}_1)
    \end{equation}
\end{itemize}

\noindent
\textbf{Optimization Objective.} As a whole, we adopt a multi-task loss function that jointly supervises mask prediction, edge localization, and mask quality estimation:
\begin{equation}
\mathcal{L} = \sum_{i \in \{1,2,4\}} \left( \mathcal{L}_{\text{mask}}^i + \mathcal{L}_{\text{edge}}^i \right) + \lambda_{\text{iou}} \mathcal{L}_{\text{iou}}
\end{equation}

\begin{table*}[ht]
\centering
%ST-Storage Tank, SP-Swimming Pool,
\setlength{\tabcolsep}{4pt}%6.5pt 
\begin{tabular}{l | l l | c c c c c c c c}
\toprule
% Method & GFLOPs & Params & Ship  & ST & LV & SV & HC & SP & Plane & mIoU \\
Method & GFLOPs & Params & Ship  & Storage Tank & LV & SV & Helicopter & Swimming Pool & Plane & mIoU \\
\midrule
% Method & GFLOPs & Params & Ship  & Storage Tank & Large Vehicle & Small Vehicle & Helicopter & Swimming Pool & Plane & mIoU \\
% SAM     &  73.02 & 74.00 & 71.11 & 60.28 & 56.85 & 64.19 & 64.29 & 66.25 \\
% SAM2    &  73.20 & 72.51 & 70.32 & 59.60 & 57.29 & 65.87 & 63.71 & 66.07 \\
% ROS-SAM &  81.61 & 81.41 & 78.94 & 72.35 & 61.31 & 79.10 & 76.52 & 75.89 \\
% UGBS    &  84.82 & 86.85 & 85.06 & 81.77 & 66.04 & 82.69 & 78.83 & 80.87 \\
% \textbf{SOPSeg}  &  \textbf{87.14} & \textbf{88.54} & \textbf{87.23} & \textbf{85.28} & \textbf{67.55} & \textbf{84.34} & \textbf{80.63} & \textbf{82.96} \\
SAM     & 1342 & 308M   & 73.02 & 74.00 & 71.11 & 60.28 & 56.85 & 64.19 & 64.29 & 66.25 \\
SAM2    & \textbf{840}  & 216.9M & 73.20 & 72.51 & 70.32 & 59.60 & 57.29 & 65.87 & 63.71 & 66.07 \\
ROS-SAM & 1594 & 359.7M & 81.61 & 81.41 & 78.94 & 72.35 & 61.31 & 79.10 & 76.52 & 75.89 \\
UGBS    & 2172 & \textbf{79.4M}  & 84.82 & 86.85 & 85.06 & 81.77 & 66.04 & 82.69 & 78.83 & 80.87 \\
\textbf{SOPSeg} & 1244 & 311M & \textbf{87.14} & \textbf{88.54} & \textbf{87.23} & \textbf{85.28} & \textbf{67.55} & \textbf{84.34} & \textbf{80.63} & \textbf{82.96} \\
\bottomrule
\end{tabular}
\caption{Comparison of IoU (\%) for different methods on iSAID dataset. Notes: LV-Large Vehicle, SV-Small Vehicle}
\label{comparision-isaid}
\end{table*}
Here, both $\mathcal{L}_{\text{mask}}^i$ and $\mathcal{L}_{\text{edge}}^i$ are composed of a sum of Binary Cross-Entropy (BCE) and DICE \cite{milletari2016v} losses between the predicted outputs and the ground truth at scale $i$. The ground-truth edge map is derived from the binary mask annotations and smoothed using a $3 \times 3$ Gaussian filter to mitigate aliasing artifacts.

The term $\mathcal{L}_{\text{iou}}$ employs a Smooth L1 loss between the predicted IoU score and the actual IoU computed from the original-resolution mask, guiding the model to produce accurate mask quality estimations.
The hyperparameter $\lambda_{\text{iou}}$ balances the contributions of different loss components. In our implementation, we set $\lambda_{\text{iou}} = 5.0$.
% The enhanced decoder complements our magnification strategy. While magnification provides higher resolution inputs, the decoder ensures these details are preserved through decoding. Edge prediction captures boundaries that might otherwise blur during upsampling, while progressive refinement leverages the increased pixel information from amplified regions. This combination enables accurate segmentation of objects as small as 10 pixels, a significant improvement over the standard SAM decoder.

\begin{figure*}[h]
\centering
\includegraphics[width=0.75\textwidth]{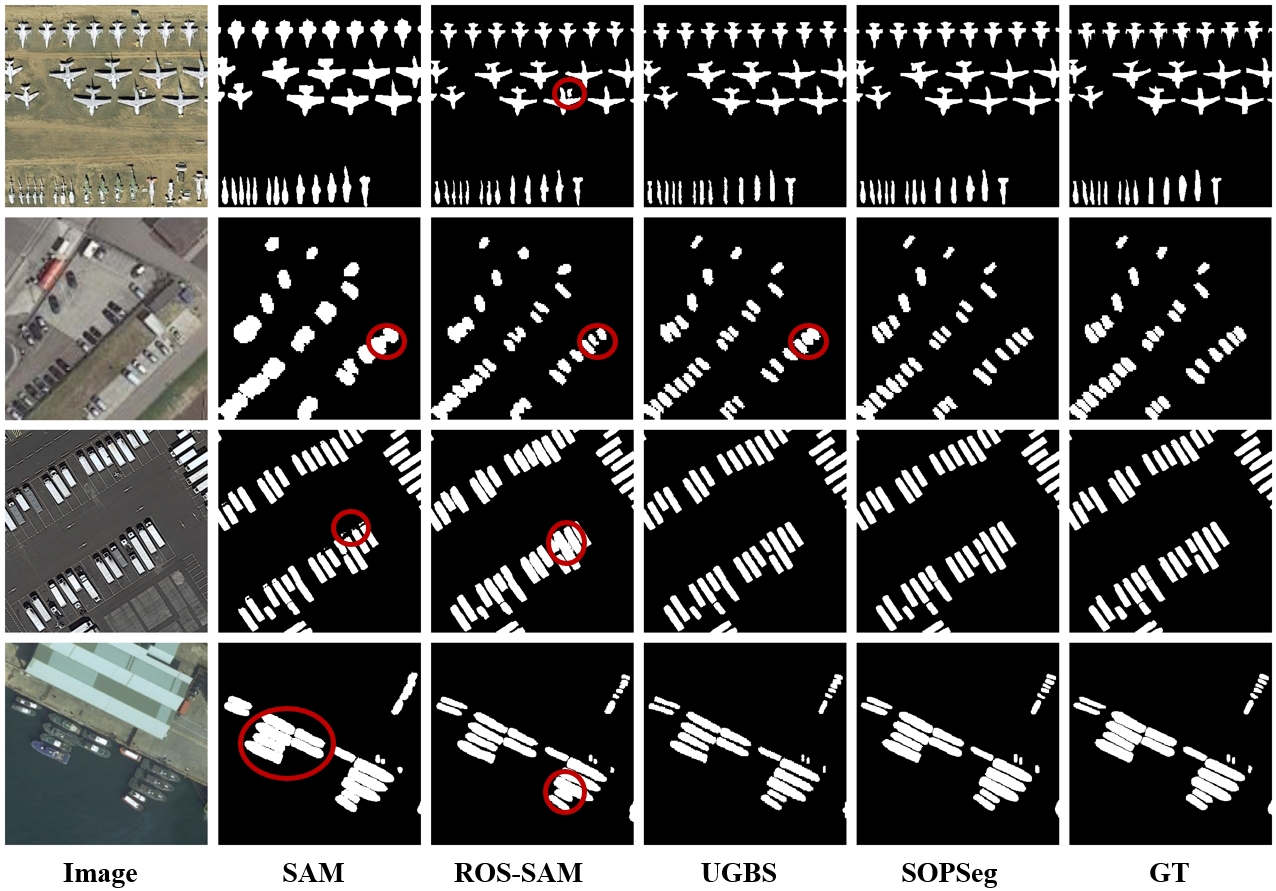}
\caption{Visualization results of small object segmentation methods on iSAID dataset.}
\label{fig:visual}
\end{figure*}

\section{Experiments}
\subsection{Experimental Setup}
\textbf{Datasets.}
We conduct prompted instance segmentation experiments on three representative remote sensing datasets: iSAID \cite{Zamir2019isaid}, NWPU-VHR10 \cite{su2019object}, and SAT-MTB \cite{li2023multitask}, all of which provide instance-level annotations for aerial or satellite imagery. The iSAID dataset is used for both training and evaluation, while NWPU-VHR10 and SAT-MTB serve as benchmarks for assessing the generalization ability of our method.

To focus on typical small object categories prevalent in remote sensing scenarios, we select specific classes from each dataset. For iSAID, we include: ship, plane, helicopter, small vehicle, large vehicle, storage tank, and swimming pool. For NWPU-VHR10, we consider: ship, airplane, vehicle, and storage tank. For SAT-MTB, we evaluate on ship and plane.

\noindent
\textbf{Evaluation Metrics.}
%各实例平均IoU, boundary IoU (BIoU),由于小目标，BIoU dilation_ratio=0.005。
We evaluate segmentation performance using mean Intersection over Union (mIoU) and boundary IoU (BIoU) \cite{cheng2021boundary} across all instances. BIoU specifically measures boundary quality by focusing on pixels near object edges. We set the dilation ratio of 0.005 to account for the limited spatial extent of small objects.

\noindent
\textbf{Comparison Method.}
We compare \textbf{SOPSeg} with several representative prompt-based segmentation methods. \textbf{SAM} \cite{kirillov2023segment} is included as the foundational model for interactive segmentation, while \textbf{SAM2} \cite{ravi2024sam2segmentimages}, which extends SAM to video segmentation, also provides more efficient image segmentation. To evaluate domain-specific adaptations to remote sensing imagery, we include \textbf{ROS-SAM} \cite{shan2025ros}. For CNN-based prompt segmentation method in remote sensing, we compare with \textbf{UGBS} \cite{yang2024Exploring}. Since UGBS relies on one instance region as input, we adopt our proposed RAM strategy to extract the surrounding regions, enabling a fair evaluation. In addition, we compare detail enhancement methods such as \textbf{MatAnything} \cite{Li2024matting} and \textbf{HQ-SAM} \cite{ke2023hqsegment} to analyze the effectiveness of their decoders relative to ours.

\noindent
\textbf{Implementation Details.} 
% SOPSeg is implemented using PyTorch and trained on NVIDIA 3090Ti GPUs. 
We use the image encoder and prompt encoder from SAM-Large and freeze their parameters during training. The enhanced decoder, initialized from SAM-Large weights, trained with learning rates of $5\times10^{-5}$. The progressive refinement module is trained from scratch with learning rates of $1\times10^{-3}$. The model is trained for 32 epochs using the AdamW optimizer and a cosine annealing learning rate schedule, with a batch size of 24.

\subsection{Comparison with Other Methods}
Table \ref{comparision-isaid} compares SOPSeg with existing prompted segmentation methods on the iSAID dataset. The results show that SOPSeg outperforms all other methods across all categories. For small vehicle class SOPSeg achieves an IoU of 85.28\%, surpassing UGBS by 3.5\% and SAM2 by 25.7\%. In addition to accuracy, we compare the computational complexity (GFLOPs on 10 instance on a image) and model sizes. SOPSeg achieves the highest mIoU (82.96\%) while maintaining a moderate GFLOPs (1244) and parameter size (311M), introducing only 3M additional parameters upon the SAM backbone..

The figure \ref{fig:visual} shows visualization results of small object prompt-based segmentation on the iSAID dataset, with all instances overlaid. Due to aggressive feature downsampling, SAM suffers from object adhesion, especially on small targets like cars. ROS-SAM and UGBS partially alleviate this issue but still struggle with boundary precision and object separation. In the first row, ROS-SAM also incorrectly segments non-aircraft regions. Our method accurately preserves object shapes across various scenes, with clear boundaries and well-separated instances, achieving results closest to the ground truth.

\noindent
\textbf{Generalization testing.} Table \ref{generalize-nwpu-mtb} evaluates method generalization on NWPU-VHR10 and SAT-MTB datasets. SOPSeg consistently outperforms all baselines. Compared with the strongest baseline UGBS, SOPSeg achieves a slight improvement on NWPU (+0.42\% IoU, +1.16\% BIoU), but a significantly larger gain on SAT-MTB (+3.06\% IoU, +3.02\% BIoU). One possible reason is that SAT-MTB presents more significant distribution differences from iSAID (trained dataset), which may affect UGBS more severely, while our SAM-based architecture remains more robust to such variations.

\begin{table}[htbp]
\centering
\begin{tabular}{lcc|cc}
\toprule
\multirow{2}{*}{Method} & \multicolumn{2}{c|}{NWPU} & \multicolumn{2}{c}{SAT-MTB} \\
 & IoU & BIoU & IoU & BIoU \\
\midrule
SAM    & 77.80 & 67.71 & 53.68 & 51.82 \\
SAM2   & 77.24 & 65.54 & 52.35 & 48.76 \\
ROS-SAM & 82.84 & 75.09 & 68.43 & 67.26 \\
UGBS    & 86.13 & 79.33 & 70.32 & 69.54 \\
\textbf{SOPSeg}  & \textbf{86.55}	& \textbf{80.49}	& \textbf{73.38}	& \textbf{72.56} \\

\bottomrule
\end{tabular}
\caption{Generalization results on NWPU and SAT-MTB datasets.}
\label{generalize-nwpu-mtb}
\end{table}

\subsection{Ablation Study}
\textbf{Effectiveness of Different Modules.}
Table~\ref{ablation} presents the ablation results on the iSAID dataset. We progressively incorporate each component of the SOPSeg framework into a baseline model, which fine-tunes the original SAM decoder using horizontal box prompts.
Adding the Region-Adaptive Magnification (RAM) module improves performance by 7.84\% mIoU. The RAM module benefits small vehicle and storage tank, where spatial details are often lost during standard downsampling.
The oriented prompt mechanism adds 2.29\% over the RAM-only configuration.
Finally, our enhanced decoder contributes an additional 1.27\% improvement. This gain is more evident for classes like plane and helicopter, which exhibit complex boundaries and fine structural details. These results demonstrate that each module brings consistent performance gains, and their combination yields the best overall segmentation performance.

\noindent
\textbf{Decoder Component Analysis.}
As shown in Table \ref{decoder-abl}, we evaluate different decoder designs by replacing our decoder with various alternatives. Our enhanced decoder achieves the best overall performance, outperforming the original SAM decoder by 1.44\% in BIoU and 1.21\% in IoU. These results demonstrate that incorporating edge prediction effectively preserves fine-grained details and improves boundary accuracy for small object segmentation.
\begin{table}[h]
\centering
\begin{tabular}{lcc}
\toprule
Method        & IoU   & BIoU  \\
\midrule
SAM          & 84.17 & 80.54 \\
MatAnything   & 84.74 & 81.02 \\
HQ-SAM        & 85.06 & 81.62 \\
\textbf{Our Decoder}   & \textbf{85.38} & \textbf{81.98} \\
\bottomrule
\end{tabular}
\caption{Evaluation on different decoder architectures using both standard IoU and boundary IoU (BIoU) metrics.}
\label{decoder-abl}
\end{table}

\begin{figure}[h]
\centering
\includegraphics[width=0.47\textwidth]{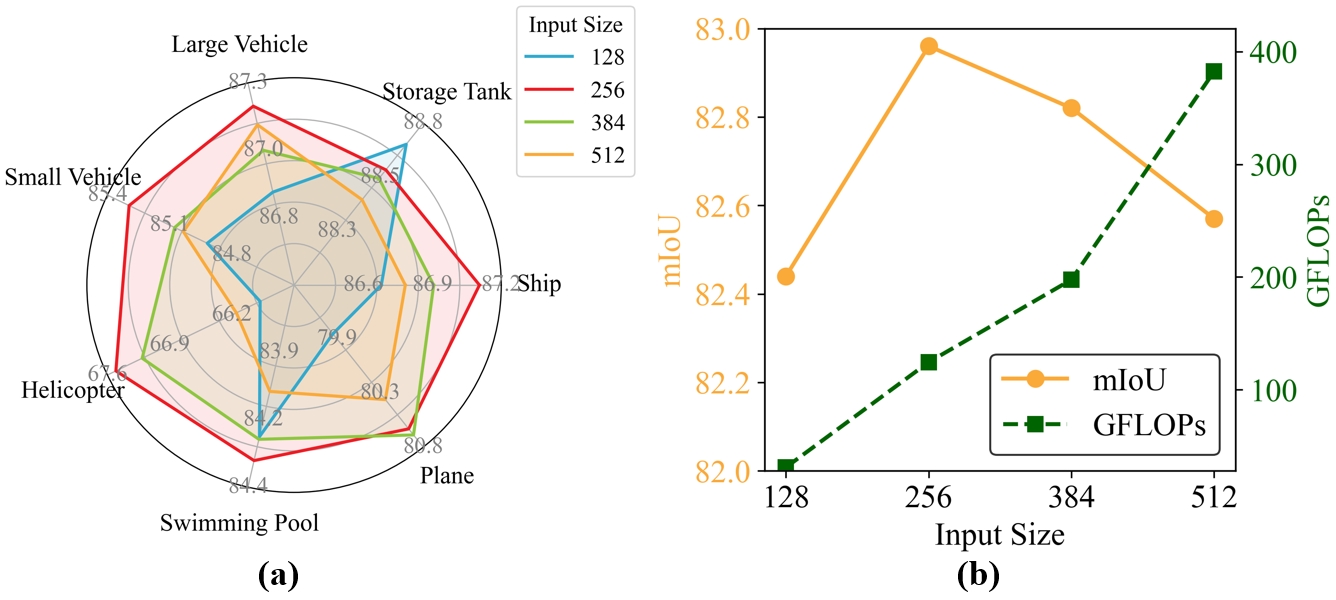}
\caption{The impact of input size on (a) class-wise IoU and (b) mean class IoU and computational cost.}
\label{fig:input_size}
\end{figure}

\noindent
\textbf{Input Resolution Analysis.}
Figure~\ref{fig:input_size} analyzes the impact of different input sizes on both segmentation performance and computational efficiency. We evaluate four input resolutions—128, 256, 384, and 512—to determine the optimal configuration for small object segmentation.

Figure~\ref{fig:input_size}(a) shows the per-category performance across different input sizes. An input resolution of 256 achieves the best overall performance across most categories, with particularly strong results for ship, swimming pool, and small vehicle. In contrast, an input resolution of 128 significantly degrades performance for categories like helicopter and plane, which rely heavily on fine-grained spatial details for precise boundary delineation. Interestingly, storage tank performs
\begin{table*}[t]
\centering
\setlength{\tabcolsep}{4pt}
\begin{tabular}{lcccccccc}
\toprule
Method    & Ship  & Storage Tank & Large Vehicle & Small Vehicle & Helicopter & Swimming Pool & Plane & mIoU \\
\midrule
Base             & 79.86 & 74.62        & 75.53          & 62.14          & 61.52      & 72.77          & 75.13 & 71.65 \\
+RAM             & 83.43 & 85.90        & 83.08          & 79.46          & 64.91      & 80.45          & 78.59 & 79.40 \\
+RAM+OPM         & 85.64 & 88.17        & 86.02          & 83.98          & 65.33      & 84.22          & 78.47 & 81.69 \\
\textbf{+RAM+OPM+EDE} & \textbf{87.14} & \textbf{88.54} & \textbf{87.23} & \textbf{85.28} & \textbf{67.55} & \textbf{84.34} & \textbf{80.63} & \textbf{82.96} \\
\bottomrule
\end{tabular}
\caption{Ablation study results on the iSAID dataset. RAM: Region-Adaptive Magnification; OPM: Oriented Prompt Mechanism; EDE: Enhanced Decoder with Edge Prediction.}
\label{ablation}
\end{table*}

\begin{table*}[t]
\centering
\setlength{\tabcolsep}{4pt}
\begin{tabular}{lccccccccc}
\toprule
Method & Plane & Helicopter & Small Vehicle & Large Vehicle & Ship & Container & Storage Tank & Swimming Pool & AP \\
\midrule
SparseInst  & 8.6  & 0.1 & 6.6  & 8.3  & 8.7  & 10.8 & 26.7 & 36.9 & 13.3 \\
Mask2Former & 25.0 & 2.2 & 8.4  & 12.1 & 13.6 & 15.2 & 33.5 & 23.8 & 16.7 \\
MaskDINO    & 41.0 & 13.1 & 22.9 & 35 & 36.7  & 40.1 & 46.5 & 45.6 & 35.1 \\
\bottomrule
\end{tabular}
\caption{Comparison of instance segmentation results (AP, \%) on our constructed  dataset.}
\label{tab:constructed_dataset}
\end{table*}

\noindent %接上段 Interestingly, storage tank performs
best at 128 resolution, possibly due to its inherently simple and compact structure, which may become over-smoothed or misrepresented when additional detail is introduced at higher resolutions.
Larger input sizes, such as 384 and 512, do not lead to proportional improvements over 256. Although marginal improvements are observed in the plane category, these come at the cost of significantly higher computational demands.

Figure~\ref{fig:input_size}(b) presents the computational analysis. The mIoU curve peaks at a resolution of 256, confirming it as the optimal choice in terms of accuracy. Meanwhile, GFLOPs (for one instance) increase dramatically with higher resolution.

\subsection{Evaluation on the Constructed Dataset}
Leveraging the strong small object segmentation capability of the proposed SOPSeg, combined with manual filtering, we constructed the \textbf{ReSOS} (\textbf{Re}mote Sensing \textbf{S}mall \textbf{O}bject \textbf{S}egmentation) dataset based on images and oriented bounding boxes from SODA-A \cite{gong2023soda}, containing pixel-level annotations for over 709k instances. This dataset provides a solid foundation for the evaluation and advancement of small object segmentation techniques. Due to page constraints, a detailed description of the dataset is provided in the appendix.

We evaluated three representative instance segmentation methods on ReSOS, with the Average Precision (AP) results across eight object categories summarized in Table \ref{tab:constructed_dataset}. MaskDINO \cite{li2023mask} achieves the highest overall AP, significantly outperforming Mask2Former \cite{cheng2022masked} and SparseInst \cite{cheng2022sparse}. The results reveal notable performance variations across categories. Storage-tank and swimming-pool classes demonstrate relatively better performance, likely due to their distinctive geometric shapes and clear boundaries that facilitate segmentation. In contrast, the helicopter category is the most challenging, primarily because of its fuzzy boundaries. Small-vehicle segmentation proves especially difficult for conventional methods, indicating that sparse representation alone is insufficient for effectively handling small objects. Moreover, we observe a consistent trend that larger objects tend to achieve higher segmentation accuracy. These findings underscore the urgent need to advance small object instance segmentation techniques.

\section{Conclusion}
We presented SOPSeg, a prompt-based framework for small object instance segmentation in remote sensing imagery. To address the loss of spatial details when objects occupy fewer than $32 \times 32$ pixels, we introduced a region-adaptive magnification strategy that preserves fine-grained features. The oriented prompt mechanism enables accurate segmentation of rotated objects without architectural modifications. The edge-aware decoder with progressive refinement enhances mask accuracy. Extensive experiments show that SOPSeg generalizes well across datasets. Ablation studies further validate the effectiveness of each proposed component.
We will release SOPSeg along with our annotated small object dataset ReSOS. This dataset fills a gap in the remote sensing community by providing pixel-level annotations for thousands of small object instances. We hope that the release of SOPSeg and its accompanying dataset will serve as a valuable resource for the community and foster continued research in small object segmentation within remote sensing imagery.

% \clearpage

% \section{Acknowledgments}
% AAAI is especially grateful to Peter Patel Schneider for his work in implementing the original aaai.sty file, liberally using the ideas of other style hackers, including Barbara Beeton. We also acknowledge with thanks the work of George Ferguson for his guide to using the style and BibTeX files --- which has been incorporated into this document --- and Hans Guesgen, who provided several timely modifications, as well as the many others who have, from time to time, sent in suggestions on improvements to the AAAI style. We are especially grateful to Francisco Cruz, Marc Pujol-Gonzalez, and Mico Loretan for the improvements to the Bib\TeX{} and \LaTeX{} files made in 2020.

% The preparation of the \LaTeX{} and Bib\TeX{} files that implement these instructions was supported by Schlumberger Palo Alto Research, AT\&T Bell Laboratories, Morgan Kaufmann Publishers, The Live Oak Press, LLC, and AAAI Press. Bibliography style changes were added by Sunil Issar. \verb+\+pubnote was added by J. Scott Penberthy. George Ferguson added support for printing the AAAI copyright slug. Additional changes to aaai2026.sty and aaai2026.bst have been made by Francisco Cruz and Marc Pujol-Gonzalez.

% \bigskip
% \noindent Thank you for reading these instructions carefully. We look forward to receiving your electronic files!

\bibliography{aaai2026}

\end{document}

\maketitle
\section*{Appendix A: Details of the ReSOS Dataset}
\subsection*{A.1 Dataset Construction Process}
We first cropped the original SODA-A images into 512×512 patches. Oriented bounding boxes from SODA-A were then used as box prompts for our SOPSeg model to generate instance masks. For quality control, we used the model-predicted IoU scores to assess the confidence of each instance prediction. An image patch was discarded if the lowest predicted IoU among all instances in that patch fell below a predefined threshold, set to 0.55 for airplane and helicopter, and 0.65 for all other categories. Additionally, low-quality masks identified through manual inspection were marked as ignore, following the convention in SODA-A.

\subsection*{A.2 Dataset Statistics}

Table~\ref{tab:resos_stats} summarizes the per-category statistics of the ReSOS dataset. The dataset comprises over 709k annotated instances across eight object categories, with a strong emphasis on small objects. We report both the mean instance area (in pixel squared) and the mean absolute size, defined as the larger dimension (height or width) of the instance's bounding box. Across categories, the mean area remains below 800~px$^2$, confirming the small-object nature of the dataset. Categories such as \textit{Swimming Pool}, \textit{Large Vehicle}, and \textit{Container} have relatively larger average sizes, likely due to their physical scale and less crowded spatial distributions. In contrast, \textit{Small Vehicle} instances are both the most abundant and the smallest in absolute scale (mean max dimension of 17.3 pixels), posing significant challenges for instance segmentation models.

\begin{table}[h]
\centering
\caption{Per-category statistics of the ReSOS dataset. The \textit{Mean Absolute Size} is the average of $\max$(height, width) per instance.}
\label{tab:resos_stats}
\begin{tabularx}{\linewidth}{lXXX}
\toprule
Category & Instance Number & Mean Area (px$^2$) & Mean Absolute Size (px) \\
\midrule
Small Vehicle  & 458,105 & 143.8 & 17.3 \\
Container      & 121,895 & 453.9 & 41.9 \\
Ship           & 60,685  & 277.4 & 25.9 \\
Storage Tank   & 35,163  & 294.6 & 18.3 \\
Swimming Pool  & 29,500  & 729.0 & 37.5 \\
Airplane       & 28,311  & 278.5 & 27.7 \\
Large Vehicle  & 15,947  & 560.0 & 46.4 \\
Helicopter     & 1,421   & 409.4 & 33.2 \\
\bottomrule
\end{tabularx}
\end{table}

\subsection*{A.3 Visualization Examples}
\begin{figure*}[htbp]
    \centering
    \includegraphics[width=\textwidth]{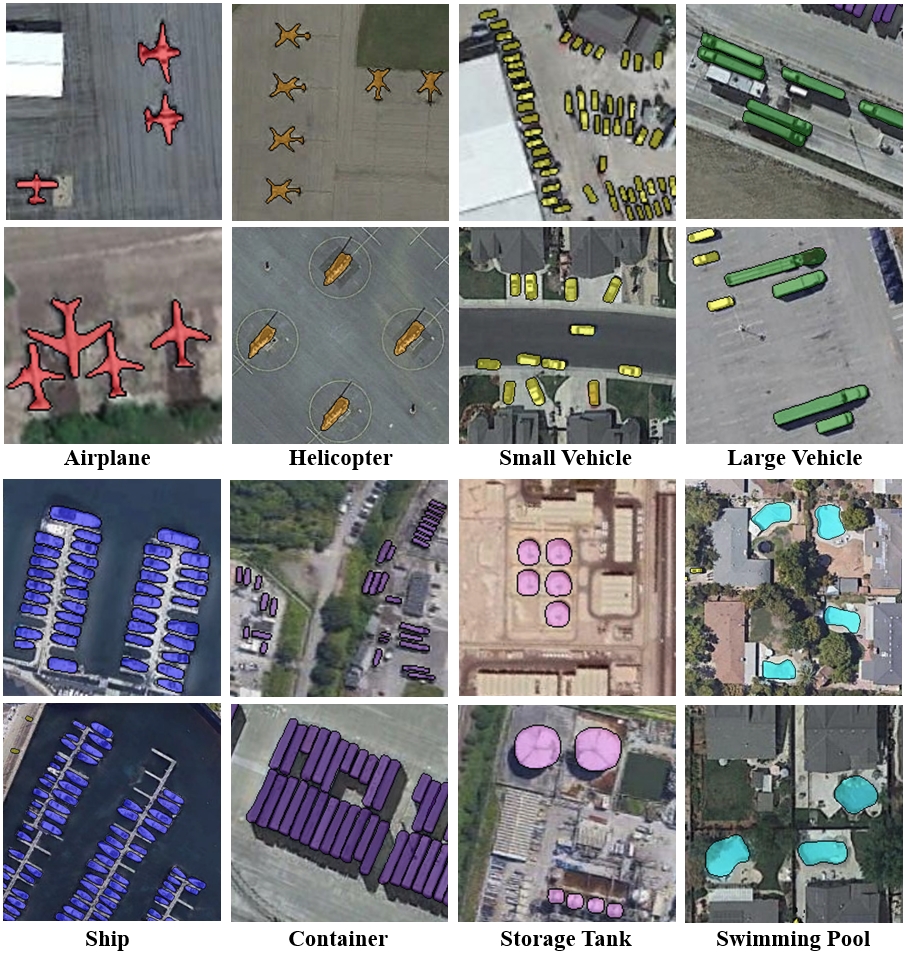}
    % \caption{visual}
    \caption{Visualization examples from the ReSOS dataset across eight object categories. Each row shows multiple sample patches with colored instance mask.}
    \label{fig:app1}
\end{figure*}
Fig \ref{fig:app1} show visualization examples from the ReSOS dataset across eight object categories. Each row represents a different object class, with multiple sample patches showing various instances within image crops. The colored masks highlight the precise boundaries of each object instance, demonstrating the dataset's annotation quality and the diversity of object appearances across different scenarios.

\section*{Appendix B: Visualization of Instance Segmentation Results on ReSOS}
\begin{figure*}[htbp]
    \centering
    \includegraphics[width=\textwidth]{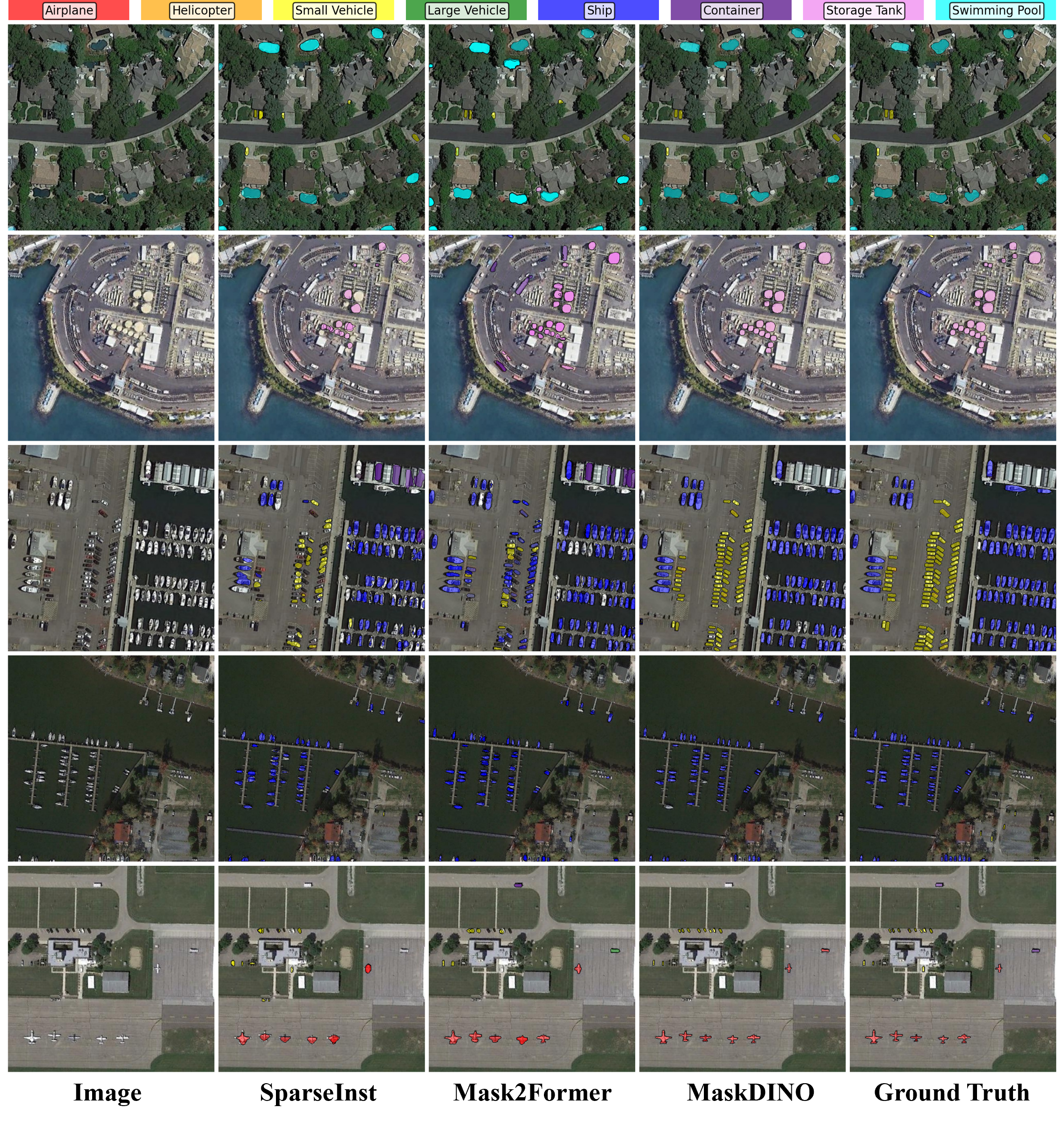}
    \caption{Comparative instance segmentation results on the ReSOS dataset. The ground truth annotations were generated by our SOPSeg.}
    \label{fig:app2}
\end{figure*}
Fig \ref{fig:app2} presents comparative instance segmentation results on the ReSOS dataset. The visualization compares performance across different methods including SparseInst, Mask2Former, MaskDINO, and ground truth annotations, which generated by our SOPSeg. The colored segmentation masks reveal how each method performs on small object instance segmentation. This comparison highlights the varying capabilities of existing methods when handling dense small objects in complex remote sensing environments, particularly for challenging categories like vehicles and aircraft.